\pdfoutput=1

\documentclass[11pt]{article}

\usepackage[preprint]{acl}

\usepackage{times}
\usepackage{latexsym}

\usepackage[T1]{fontenc}

\usepackage[utf8]{inputenc}

\usepackage{microtype}

\usepackage{inconsolata}

\usepackage{graphicx}

\usepackage[utf8]{inputenc} 
\usepackage[T1]{fontenc}    
\usepackage{hyperref}       
\usepackage{url}            
\usepackage{booktabs}       
\usepackage{amsfonts}       
\usepackage{nicefrac}       
\usepackage{microtype}      
\usepackage{xcolor}         

\usepackage{multirow}
\usepackage{graphicx}
\usepackage{ulem}
\usepackage{caption}
\usepackage{subcaption}
\usepackage{amssymb}
\usepackage{booktabs}
\usepackage{multibib}
\usepackage{colortbl}
\usepackage{makecell}
\usepackage{wrapfig}
\usepackage{color}
\usepackage{float}
\usepackage{inconsolata}
\usepackage{amsmath}

\usepackage{enumitem}
\setlist{leftmargin=1em}

\usepackage[listings]{tcolorbox}
\tcbuselibrary{listings,theorems}
\newtcolorbox{mybox}{colback=white!5!white,colframe=black!75!black, left=.05in, right=.05in}

\definecolor{bluex}{rgb}{0.27, 0.42, 0.81}
\definecolor{purplex}{HTML}{9564bf}
\definecolor{red3}{HTML}{C52A20}
\definecolor{red2}{HTML}{B36A6F}
\definecolor{red1}{HTML}{FFb5b5}
\definecolor{purple}{HTML}{B36A6F}
\definecolor{darkyellow}{HTML}{D5BA82}
\definecolor{blue1}{HTML}{508AB2}
\definecolor{blue2}{HTML}{C4E4E3}
\definecolor{green1}{HTML}{A1D0C7}
\definecolor{green2}{HTML}{BFF6BA}
\definecolor{green3}{HTML}{028100}
\definecolor{teal}{HTML}{508AB2}
\definecolor{purple1}{HTML}{8d3a94}

\newtcbtheorem[]{exmp}{Example}%
{colback=purple1!5,colframe=purple1!80,fonttitle=\bfseries, left=.02in, right=.02in,bottom=.02in, top=.02in}{exmp}

\DeclareMathOperator*{\argmax}{arg\,max}
\DeclareMathOperator*{\argmin}{arg\,min}

\title{Dehallucinating Parallel Context Extension for Retrieval-Augmented Generation}

\author{Zexiong Ma\thanks{\, Work done during the internship at Microsoft.}\hspace{0.4mm} $^{\diamondsuit,\clubsuit}$,\, Shengnan An$^{*\heartsuit,\clubsuit}$,\, Zeqi Lin\thanks{\, Corresponding authors.}\hspace{0.4mm} $^{\clubsuit}$,\\\textbf{Yanzhen Zou}$^{\diamondsuit}$,\, \textbf{Jian-Guang Lou}$^{\clubsuit}$,\, \textbf{Bing Xie}$^{\dagger\diamondsuit}$
\vspace{1mm}\\
  $^{\diamondsuit}$School of Computer Science, Peking University,\,\,
  $^{\clubsuit}$Microsoft,\, 
  $^{\heartsuit}$Xi’an Jiaotong University\vspace{1mm}\\
  $^{\diamondsuit}$\texttt{mazexiong@stu.pku.edu.cn, \{zouyz, xiebing\}@pku.edu.cn}, \\
  $^{\heartsuit}$\texttt{an1006634493@stu.xjtu.edu.cn}, $^{\clubsuit}$\texttt{\{Zeqi.Lin, jlou\}@microsoft.com}
}

\begin{document}
\maketitle
\begin{abstract}
Large language models (LLMs) are susceptible to generating hallucinated information, despite the integration of retrieval-augmented generation (RAG).
Parallel context extension (PCE) is a line of research attempting to effectively integrating parallel (unordered) contexts, while it still suffers from in-context hallucinations when adapted to RAG scenarios.
In this paper, we propose \textbf{DePaC} (\textbf{De}hallucinating \textbf{Pa}rallel \textbf{C}ontext Extension), which alleviates the in-context hallucination problem with \textbf{context-aware negative training} and \textbf{information-calibrated aggregation}.
DePaC is designed to alleviate two types of in-context hallucination: \textbf{fact fabrication} (i.e., LLMs present claims that are not supported by the contexts) and \textbf{fact omission} (i.e., LLMs fail to present claims that can be supported by the contexts).
Specifically, (1) for fact fabrication, we apply the context-aware negative training that fine-tunes the LLMs with negative supervisions, thus explicitly guiding the LLMs to refuse to answer when contexts are not related to questions;
(2) for fact omission, we propose the information-calibrated aggregation which prioritizes context windows with higher information increment from their contexts.
The experimental results on nine RAG tasks demonstrate that DePaC significantly alleviates the two types of in-context hallucination and consistently achieves better performances on these tasks.
\end{abstract}

\section{Introduction}

Retrieval-augmented generation (RAG)~\citep{lewis2020retrieval, gao2023retrieval} is nowadays a prevalent paradigm for incorporating large language models (LLMs) ~\citep{openai2023gpt4, touvron2023llama, jiang2023mistral} with outside knowledge.
RAG employs a \textit{retriever} to fetch  documents that are semantically closest to the question, and incorporates them into LLM's prompt. 
Parallel Context Extension (PCE) ~\citep{hao2022structured, ratner2023parallel,su2024naive} is a line of research attempting to effectively integrating parallel contexts through an aggregation function. PCE is highly compatible with RAG scenarios, as the candidate retrieved documents of RAG are independ of each other.

\begin{figure*}[t]
    \centering
    \begin{subfigure}{.99\textwidth}
        \centering
       \includegraphics[width=.99\textwidth]{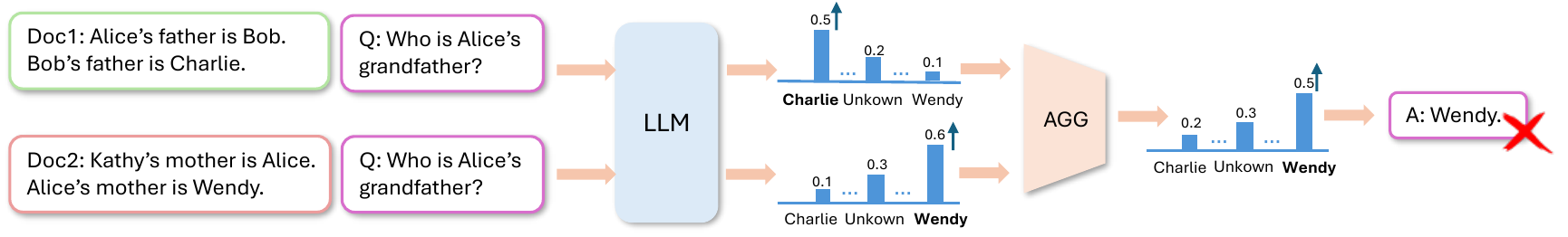}
        \caption{
        Fact fabrication example. $Doc_2$ is useless to answer the question. The higher confidence in \textit{"Wendy"} on $Doc_2$ caused PCE to fabricate the answer "Alice's grandfather is Wendy."
        \\
        }
        \label{fig:motivation1}
    \end{subfigure}
    \begin{subfigure}{.99\textwidth}
    \centering
       \includegraphics[width=.99\textwidth]{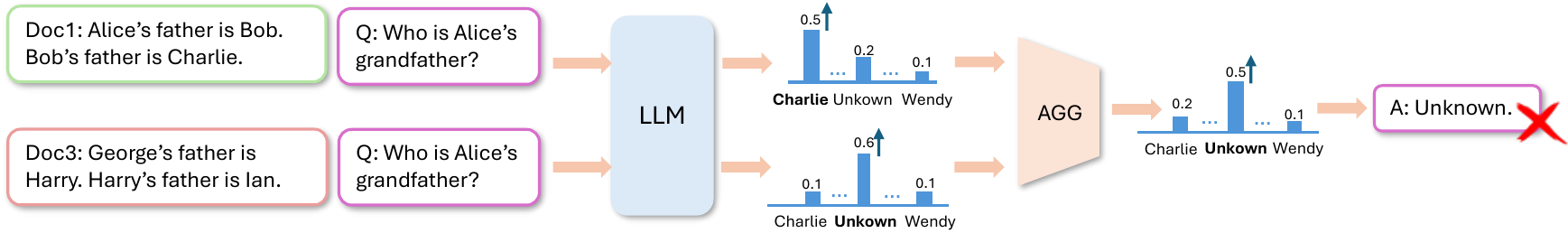}
        \caption{
        Fact omission example. $Doc_3$ is useless to answer the question. The higher confidence in "unknown" on $Doc_3$ caused PCE to omit the fact on $Doc_1$, resulting an incorrect final answer after aggregation.
        }
        \label{fig:motivation2}
    \end{subfigure}
    \caption{
        Existing PCE approaches face two types of in-context hallucination issues when applied to RAG: (1) Fact fabrication. LLM generates fabricated answers that are inconsistent with the contextual facts. (2) Fact omission. The absence of required information in certain windows disproportionately influence the aggregation function, leading to disregard critical information in other windows.
        }
        \label{fig:motivation}
\end{figure*}

However, existing PCE approaches still face two types of in-context hallucination issues~\citep{ji2023survey, rawte2023survey, yang2023revisiting}:  \textbf{fact fabrication} and \textbf{fact omission.}  (1) \textbf{fact fabrication} occurs when the model presents fabricated claims that are inconsistent with the contextual facts. As shown in Figure \ref{fig:motivation1}, LLM confidently produces a fabricated answer for the window with $Doc_2$, caused PCE to fabricate the wrong answer. (2) \textbf{fact omission} refers to windows lacking useful information may disproportionately affect the aggregation function, leading it to omit critical information present in other windows. This will make LLMs fail to present claims that can be supported by the contexts. As shown in Figure \ref{fig:motivation2}, $Doc_3$ does not contain required information, makes LLM confidently generate \textit{"Unknown"} for the window with $Doc_3$, further leading to the wrong final answer.

\begin{figure}[t]
\vspace{-5mm}
\centering
\includegraphics[width=0.8\columnwidth]{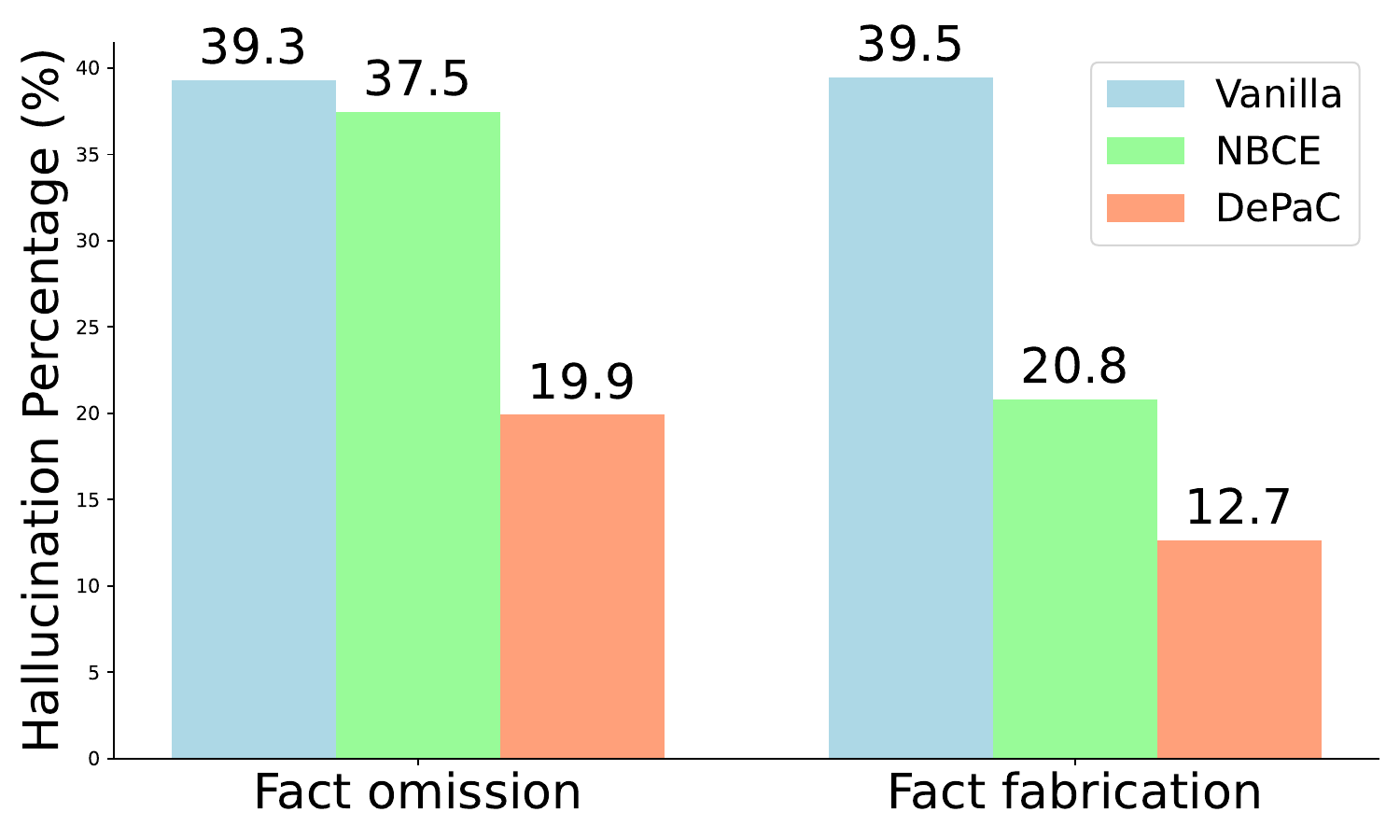}
\caption{DePaC significantly reduces the occurrence of hallucinations in responses within RAG scenarios.}
\label{fig:hallucination_percentage_avg}
\vspace{-3mm}
\end{figure}

In this paper, we propose DePaC to alleviate the hallucination issue of parallel context extension on RAG. DePaC contains two parts:  \textbf{NegTrain} (Context-aware \textbf{Neg}ative \textbf{Train}ing) to address fact fabrication issue and \textbf{ICA} (\textbf{I}nformation-\textbf{C}alibrated \textbf{A}ggregation) to address fact omission issue.
(1) \textbf{NegTrain} guids the LLMs to refuse to answer when contexts are not related to the question. NegTrain consists of two parts of training data: one part comprises useful documents and questions as input, with corresponding answers as output. While the other part treats irrelevant documents and questions as input, with a rejection token as output.  (2) \textbf{ICA} prioritizes context windows with higher information increment from their contexts. Specifically, we utilize Kullback-Leibler~\citep{10.1214/aoms/1177729694} divergence to measure the information increment of with-document compared to non-document.
This approach enhances DePaC's capability to identify useful information within parallel windows.
Moreover, DePaC has lower computational complexity than vanilla inference approach. The inference time of DePaC increases linearly with the number of documents, while inference time of vanilla approach increases squarely.

We conduct experiments on various RAG tasks, demonstrate that DePaC significantly alleviates the two types of hallucination and consistently achieves promising performances.
Then we analyze the proportion of hallucination produced by different approaches, demonstrating that DePaC can effectively mitigate the two types of hallucination (Figure \ref{fig:hallucination_percentage_avg}).
We also conducte ablation study to identify that information-calibrated aggregation and context-aware negative training are both essential for DePaC performance. 

The main contents of this paper are organized as follows. Section \ref{sec:background} introduces the formalization of PCE and two existing aggregation methods for PCE.
Section \ref{sec:DePaC} introduces the methodology and implementation details of DePaC.
Section \ref{sec:complexity} introduces the complexity analysis of DePaC.
Section \ref{sec:setup} introduces our experimental results on information seeking and DocQA.
Section \ref{sec:related} discusses the related work.
Finally, section \ref{sec:conclusion} provides a conclusion regarding our work.

\section{Background: Parallel Context Extension (PCE)}\label{sec:background}


The core idea of PCE involves aggregating information from multiple context windows into a unified representation space.
Such a representation aggregation can be formalized on either the probability distributions of output tokens~\citep{su2024naive}, or the internal hidden states in attention layers~\citep{hao2022structured, ratner2023parallel}. 
~\citet{su2024naive} claimed the above two formalizations have similar practical performances.
In this work, we adopt the formalization in~\citep{su2024naive} that takes the aggregation of output distributions.

Given an question $\mathcal{Q}$, a set of retrieved documents $\mathcal{D}=\{d_1,d_2,...,d_n\}$, and a language model with parameters $\theta$, PCE first computes the output distribution of each context window,
\begin{equation}
    \mathbf{p_{i,j}} = p_{\theta}(\ \cdot\ |\ d_{j} \oplus \mathcal{Q} \oplus \mathcal{A}_{1:i-1}),
\end{equation}
where $\mathbf{p_{i,j}}$ is the probability distribution of the $i$-th token for output $\mathcal{A}$ based on the $d_{j}$ document, and $\oplus$ represents the concatenation of sequences.
Subsequently, these individual distributions are aggregated into a single distribution,
\begin{equation}
    \mathbf{p_{i}} = \mathrm{AGG}(\mathbf{p_{i,1}},\ \mathbf{p_{i,2}},\ ...,\ \mathbf{p_{i,n}}),
\end{equation}
where $\mathrm{AGG}(\cdot)$ represents the aggregation method.
Finally, the output token $\mathcal{A}_{i}$ will be sampled based on the aggregated distribution $\mathbf{p_{i}}$,
\begin{equation}
    \mathcal{A}_{i} \sim \hat{\mathbf{p_{i}}},\quad 
    \hat{\mathbf{p_{i}}} = \mathbf{p_{i}} - \alpha \cdot \mathbf{p_{i,c}},\quad 
\end{equation}
\begin{equation}
    \mathbf{p_{i,c}} = p_{\theta}(\ \cdot\ |\ \mathcal{Q} \oplus \mathcal{A}_{1:i-1}),
\end{equation}
where the $\hat{\mathbf{p_{i}}}$ is the calibrated distribution to facilitate generation.
We set $\alpha=0.2$ following~\citet{su2024naive}.


The effectiveness of the PCE paradigm is significantly influenced by the design of the aggregation method $\mathrm{AGG}(\cdot)$. Here, we discuss two aggregation methods used in existing studies.

\textbf{Average Aggregation}~\citep{hao2022structured, ratner2023parallel}.
The aggregated distribution is computed as the average of $n$ individual distributions,
\begin{equation}\label{equ:avg}
    \mathbf{p_{i}} = \frac{1}{n}\sum_{j=1}^{n}\mathbf{p_{i, j}}.
\end{equation}
In practice, the size of the retrieved document set $\mathcal{D}$ can be large, potentially containing only a few relevant documents. Average aggregation treats each context window with equal importance, makes it unable to seek critical information when applied to RAG.

\begin{figure*}[t]
    \centering
    \includegraphics[width=.99\textwidth]{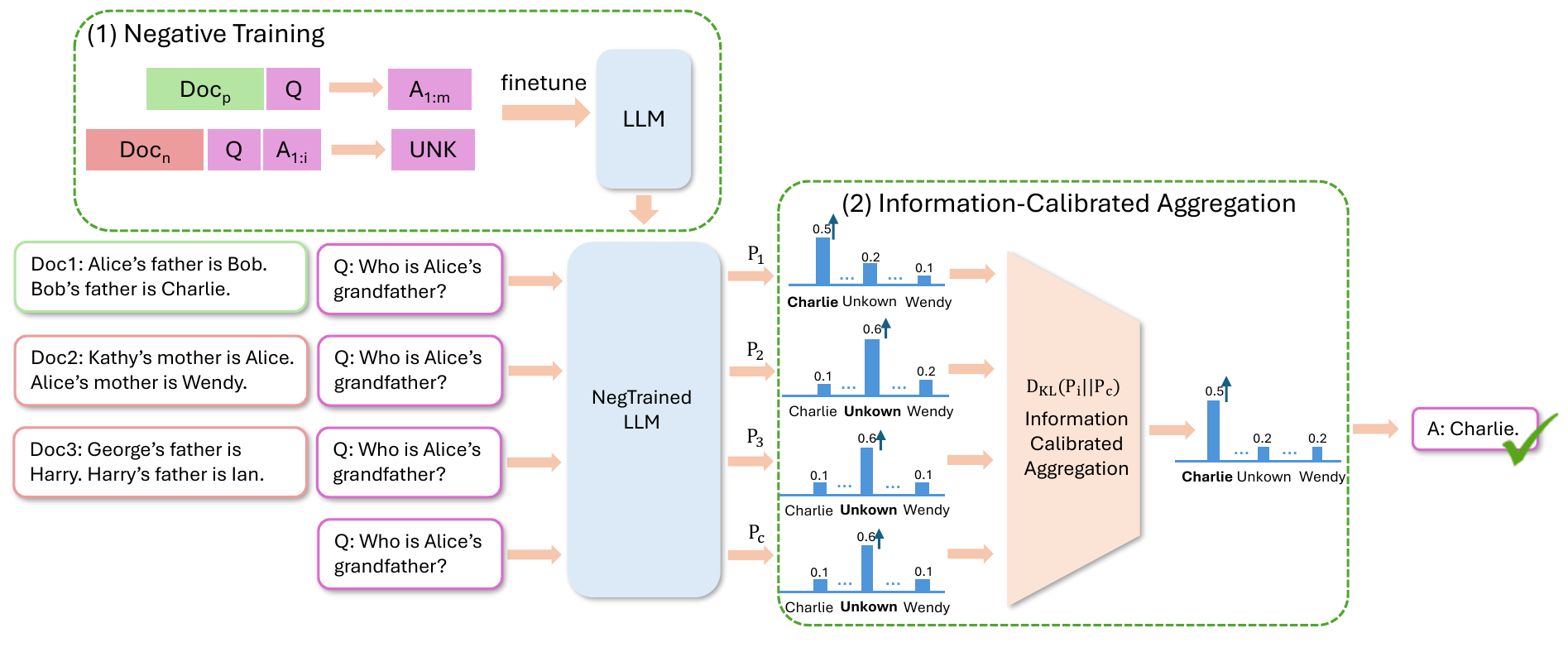}
    \caption{
    DePaC consists of two key components: (1) a context-aware negative training technique to alleviate fact fabrication, and (2) an information-calibrated aggregation method to alleviate fact omission.
    }
    \label{fig:approach}
\end{figure*}

\textbf{Lowest-Uncertainty Aggregation}~\citep{su2024naive}.
This method selects the individual distribution with the lowest uncertainty as the aggregation result,
\begin{equation}\label{equ:nbce}
    \mathbf{p_{i}} = \argmin_{\mathbf{p_{i, j}}}H(\mathbf{p_{i, j}}),\quad
\end{equation}
\begin{equation}
    H(\mathbf{p_{i, j}}) = -\mathbf{p_{i, j}} (\log\mathbf{p_{i, j}})^{T}.
\end{equation}
Lowest-uncertainty aggregation addresses the limitations of average aggregation by filtering out high-uncertainty windows. However, it remains a sub-optimal solution as it still suffers from the two types of hallucination illustrated in Figure \ref{fig:motivation}.

\section{Dehallucinating Parallel Context Extension (DePaC)} \label{sec:DePaC}



As shown in Figure \ref{fig:approach}, we propose two methods to alleviate the fact fabrication and fact omission hallucinations of PCE for RAG scenarios. First, we introduce \textbf{Context-aware Negative Training} to enable the model to refuse to answer questions when the relevant information is missing in the context, thereby mitigating fact fabrication. Then, we propose \textbf{Information-Calibrated Aggregation} to measure the information increment given by the document, preventing the model from fact omission.

\paragraph{Context-aware Negative Training (NegTrain).}
We introduce context-aware negative training to alleviate fact fabrication, which explicitly train the backbone model to determine whether a question is answerable based on the provided document. 
If not, we hope the model to refuse to answer the question rather than generating hallucinations.

Given an RAG example with a question $\mathcal{Q}$, a ground-truth answer $\mathcal{A}$, and a retrieved document $d_{j}$, we fine-tune the backbone model $\theta$ according to the following loss function,
\begin{flalign}
    &\mathrm{Loss}(\mathcal{Q},\mathcal{A}_{1:m},d_{j})=&&\\\nonumber
    &\begin{cases}
        \mathrm{CE}[p_{\theta}(\ \cdot\ |\ d_{j} \oplus \mathcal{Q} \oplus \mathcal{A}_{1:i}),\ t_{d}],& \text{$\mathcal{Q}$ unrelated to $d_{j}$,}\\
        \mathrm{CE}[p_{\theta}(\ \cdot\ |\ d_{j} \oplus \mathcal{Q}),\ \mathcal{A}_{1:m}],& \text{$\mathcal{Q}$ related to $d_{j}$,}
    \end{cases}&&
\end{flalign}
where $\mathrm{CE}[\cdot]$ represents the cross-entropy loss, $t_d$ is a pre-defined \textbf{rejection token}, $m$ refers to the sequence length of the ground-truth answer, $\mathcal{A}_{1:m}$ refers to the complete ground-truth answer with all tokens, $\mathcal{A}_{1:i}$ refers to the partial ground-truth answer the first tokens. 
As shown in Figure \ref{fig:approach}(1), to prevent DePaC from generating rejection token only at the beginning of the answer, we also include the positive answer clauses as input. 
After context-aware negative training, we use $t_d$ to explicitly judge the usefulness of each context window. 
We set $t_d$ as the \texttt{UNK} token to minimize interference with normal tokens during training.

\paragraph{Information-Calibrated Aggregation (ICA).}
As discussed in Section~\ref{sec:background}, merely measuring the uncertainty of the final output distribution can be heavily influenced by fact omission hallucination. 
We propose to measure the changes of uncertainty from the non-document output distribution to the with-document output distribution, reflecting the information increment provided by the retrieved document.

Specifically, we apply the Kullback-Leibler (KL) divergence to measure the information increment,
\begin{equation}
    \Delta(\mathbf{p_{i, j}}, \mathbf{p_{i, c}}) = \mathrm{D_{KL}}(\mathbf{p_{i, j}}\ ||\ \mathbf{p_{i, c}}),\quad
\end{equation}
\begin{equation}
    \mathbf{p_{i,c}} = p_{\theta}(\ \cdot\ |\ \mathcal{Q} \oplus \mathcal{A}_{1:i-1}),
\end{equation}
where $\mathbf{p_{i,c}}$ is the non-document output distribution.

Finally, we integrate the above two methods as two penalty terms to inject into Equation~\ref{equ:nbce},
\begin{flalign}\label{equ:ours_ori_p}
    &\mathbf{p_{i}} = &&\\\nonumber
    &\argmin_{\mathbf{p_{i, j}}} C(\mathbf{p_{i, j}}, \mathbf{p_{i, c}}) - \gamma\cdot\mathbb{I}(\argmax_{k}\mathbf{p_{i, j}}^{k}=t_{d}),\quad&&
\end{flalign}

\begin{equation}\label{equ:ours_ori_c}
    C(\mathbf{p_{i, j}}, \mathbf{p_{i, c}}) = H(\mathbf{p_{i, j}}) - \beta\cdot\Delta(\mathbf{p_{i, j}}, \mathbf{p_{i, c}}),\quad
\end{equation}

where $\mathbb{I}[\cdot]$ represents the indicator function, $\mathbf{p_{i, j}}^{k}$ is the output probability on $k$-th token in the vocabulary, and $\beta>0$ and $\gamma>0$ are hyper-parameters.
Equation~\ref{equ:ours_ori_p} and~\ref{equ:ours_ori_c} mean that the selected context window should have low uncertainty and high information increment, and should not be aligned to the rejection token. Finally, the output token $\mathcal{A}_{i}$ will be sampled based on the aggregated distribution $\mathbf{p_{i}}$. For ease of implementation, we provide a simplified form of DePaC in Appendix \ref{sec:simplified_form}.

\paragraph{Implementation Details}\label{sec:implementation}
Following previous work~\citep{an2024make}, we use the C4~\citep{raffel2020exploring} corpus to construct our context-aware negative training dataset. For a segment of text from C4, we first split it into text fragments with a maximum length of 4k tokens. 
We first sample a fragment serves as oracle document, and use GPT-4-Turbo to generate questions and answers based on the oracle document as positive training data. Then we sample unrelated fragment serves as distractor document to construct context-aware negative training data based on the positive ones. 
To prevent the model from overfitting on $t_d$, we control $t_d$ occurrence to match the average frequency of the 2,000 most frequent tokens in NegTrain. Finally, we construct 19K samples for context-aware negative training. We fine-tune Mistral-7B~\citep{jiang2023mistral} using 8x80G A100 GPUs, set the global batch size as 128 and trained for two epochs. We use Flash Attention-2~\citep{dao2023flashattention2} to enhance the training speed.  The entire training process takes about 4 hours.



\begin{figure*}[t]
    \centering
    \includegraphics[width=.49\textwidth]{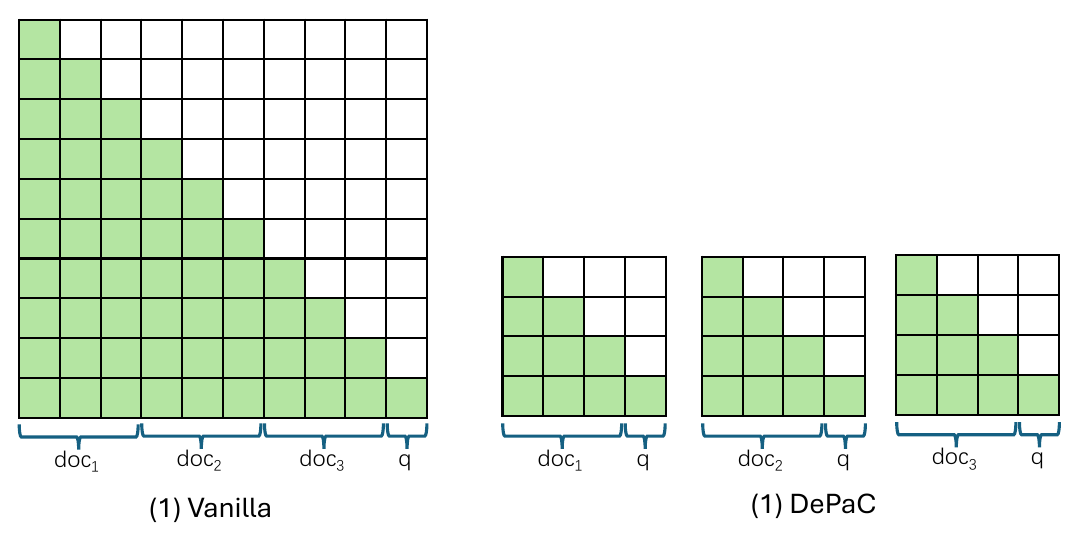}
    \includegraphics[width=.5\textwidth]{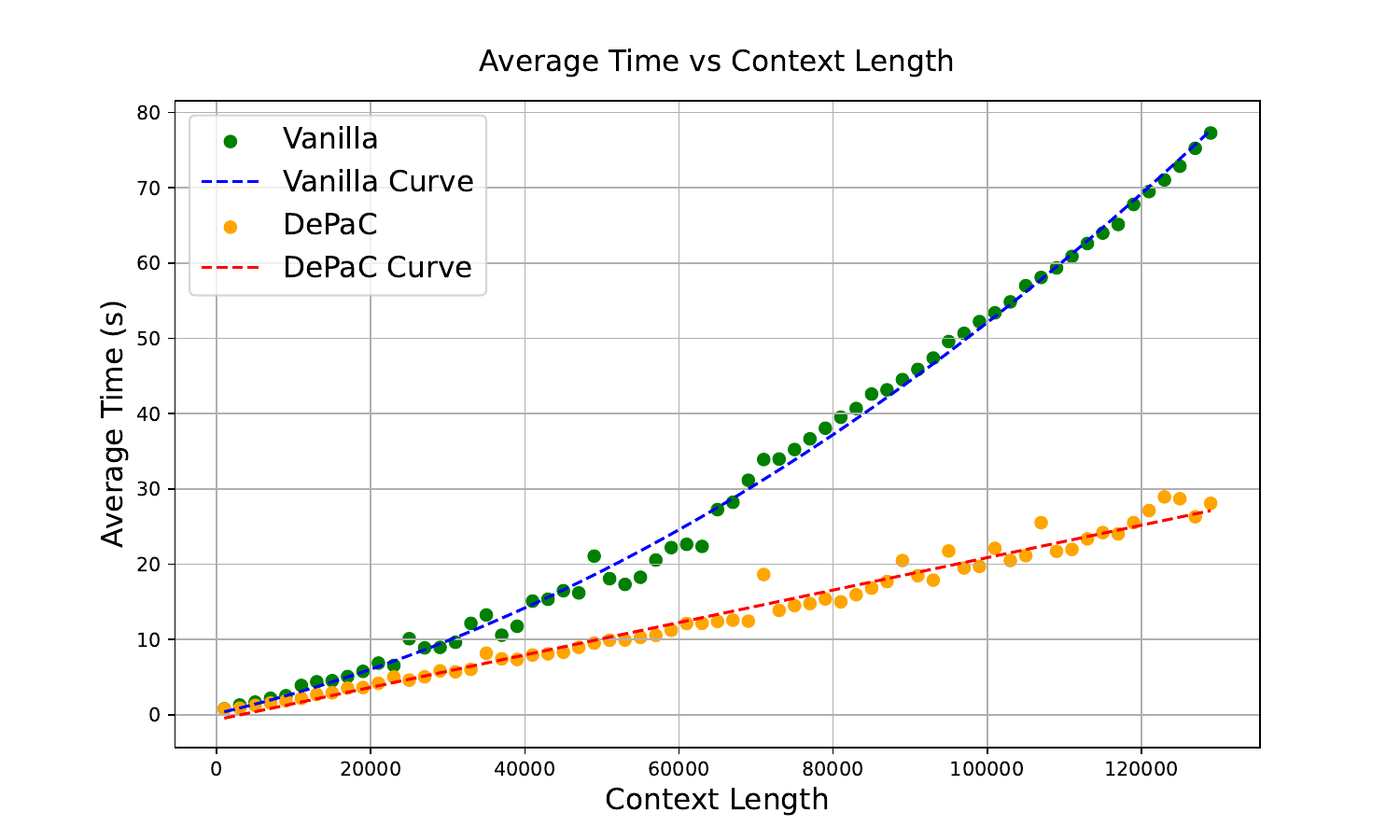}
    \caption{Attention pattern and execution time comparison between DePaC and vanilla inference. The execution time of DePaC increases linearly with context length, while vanilla's complexity grows quadratically.}
    \label{fig:attention_mask}
\end{figure*}

\section{Complexity Analysis}\label{sec:complexity}
Considering that RAG scenarios have high expectations for execution efficiency and previous PCE-style work lacked analysis of the execution efficiency, we present the inference complexity of DePaC compared with vanilla inference approach. 
Figure \ref{fig:attention_mask} shows the attention pattern and execution time comparison between DePaC and vanilla inference. As the length of the question is much smaller than the length of the document, the complexity of processing the question is ignored. 
Given a LLM with $m$ layers, we assume that the context consists of $k$ documents, each with $n$ tokens.

\paragraph{Vanilla complexity.} Vanilla inference directly concatenates the $k$ documents as the input to LLM, with a sequence length of $kn$. The attention of each layer is calculated by $\text{Attention}(Q, K, V) = \text{softmax}\left({QK^T}\right) V$, 
where $Q,K,V \in \mathbb{R}^{(kn) \times d}$ is the query, key and value matrix. The complexity of $QK^T$ is $\mathcal{O}((kn)^2 \cdot d)$. The complexity of applying softmax to the matrix is $\mathcal{O}((kn)^2)$. After applying softmax, the complexity of multiplying with $V$ is $\mathcal{O}((kn)^2 \cdot d)$. So the complexity of $Attention(Q,K,V)$ for $m$ layers is $\mathcal{O}(k^2 \cdot n^2 \cdot d \cdot m)$.

\paragraph{DePaC complexity.} In DePaC, $k$ documents are inputted to LLM in parallel, the sequence length for each input is $n$. This is akin to $k$ times $Attention(Q,K,V)$ computions, but with smaller $Q,K,V \in \mathbb{R}^{n \times d}$, so the complexity of $Attention(Q,K,V)$ for $m$ layers is $\mathcal{O}(k \cdot n^2 \cdot d \cdot m)$. The complexity of calculating KL divergence for $k$ documents is $\mathcal{O}(k \cdot n)$. Hence, the complexity of Attention compution for DePaC inference is $\mathcal{O}(k \cdot n^2 \cdot d \cdot m)$.

The complexity of Vanilla increases quadratically with $k$, while DePaC's complexity grows linearly. Figure \ref{fig:attention_mask} shows the average execution time of DePaC and vanilla inference approach with different context length, DePaC has faster inference speed than vanilla approach. Moreover, DePaC can place all documents in a single batch for parallel processing, further enhancing DePaC's inference speed.


\section{Experiments} \label{sec:setup}
We conduct experiments on various tasks to assess DePaC's performance on RAG and alleviate the two types of in-context hallucination. 


\begin{table*}[t]
\renewcommand\arraystretch{1.2}
\caption{DocQA results. We evaluete on three QA datasets with $k$=5,10,20 candidate documents.}
\label{tab:qa}
\centering
\resizebox{.99\linewidth}{!}{
\begin{tabular}{l|ccc|ccc|ccc|ccc}
\toprule
\multirow{2}{*}{Method} & \multicolumn{3}{c}{Qasper} & \multicolumn{3}{c}{MultifieldQA} & \multicolumn{3}{c}{NarrativeQA} & \multicolumn{3}{c}{Avg}\\
& $k$=5 & $k$=10 & $k$=20 & $k$=5 & $k$=10 & $k$=20 & $k$=5 & $k$=10 & $k$=20 & $k$=5 & $k$=10 & $k$=20 \\
\midrule
Vanilla~\citep{jiang2023mistral} & 15.0 & 13.3 & 8.6 & 39.7 & 33.4 & 31.6 & 10.2 & 9.1 & 9.6 & 21.6 & 18.6 & 16.6\\
YaRN~\citep{peng2023YaRN}    & 7.6 & 7.2 & 5.5 & 30.2 & 31.7 & 23.6 & 1.2 & 2.1 & 2.3 & 13.0 & 13.7 & 10.5 \\
AVP~\citep{hao2022structured}     & 6.7 & 6.6 & 6.7 & 16.7 & 15.3 & 15.4 & 8.6 & 8.5 & 8.3 & 10.7 & 10.1 & 10.1\\
NBCE~\citep{su2024naive}    & 11.7 & 9.9 & 9.8 & 31.0 & 29.0 & 26.9 & 15.9 & 15.8 & 15.1 & 19.5 & 18.2 & 17.3\\
DePaC (ours)   & \textbf{17.3} & \textbf{16.0} & \textbf{14.8} & \textbf{40.7} & \textbf{40.6} & \textbf{40.9} & \textbf{16.4} & \textbf{16.3} & \textbf{16.0} & \textbf{24.8} & \textbf{24.3} & \textbf{23.9}\\
\bottomrule
\end{tabular}
}
\end{table*}

\begin{table*}[t]
\renewcommand\arraystretch{1.2}
\caption{DocQA results of different PCE approaches with Llama3-8B. We set document number $k$=5 for all datasets.}
\label{tab:qa_llama}
\centering
\resizebox{.6\linewidth}{!}{
\begin{tabular}{lcccc}
\toprule
Method & Qasper & MultifieldQA & NarrativeQA & Avg\\
\midrule
Vanilla~\citep{jiang2023mistral} & 7.2 & 9.6 & 6.4 & 7.7 \\
AVP~\citep{hao2022structured} & 6.1 & 8.2 & 5.6 & 6.6 \\
NBCE~\citep{su2024naive} & 9.9 & 15.6 & 13.9 & 13.1 \\
DePaC (ours) & \textbf{17.6} & \textbf{41.0} & \textbf{14.1} & \textbf{24.2} \\
\bottomrule
\end{tabular}
}
\end{table*}

\subsection{Tasks}
We conduct evaluations on nine RAG tasks, including six information seeking tasks and three document-based question-answering tasks.

The \textbf{information seeking} tasks serve to explicitly probe the information awareness of DePaC.
Each test case in these tasks contains an information query question and a large amount of contexts.
Based on the given question, the model is required to seek for some textual pieces within the contexts.
The information seeking tasks include:

\begin{itemize}
    \item \textbf{Function name retrieve (FuncNR)}~\citep{an2024make}.
    The contexts in FuncNR contain a large number of Python functions, all of which are sampled from the training data of Starcoder~\citep{li2023starcoder}.
    The questions in  FuncNR ask for retrieving the function names based on the given code snippets.
    We extend the original context length in~\citet{an2024make} from 32K to 128K.
    
    \item \textbf{Entity label retrieve (EntLR)}~\citep{an2024make}.
    The contexts in EntLR contain a large number of entities, all of which are sampled from Wikidata.
    Each entity is a triplet in the form of (id, label, description).
    The questions in EntLR ask for retrieving the labels corresponding to the given entity ids from the contexts.
    We extend the original context length in~\citet{an2024make} from 32K to 128K.
    
    \item \textbf{Multi-values Needle-in-a-Haystack (MVIH)}~\citep{hsieh2024ruler}.
    The contexts in MVIH contain multiple values for a certain key, along with other unrelated text pieces.
    The questions in MVIH require the model to seek for all the associated values for the given key.

    \item \textbf{APIBench}~\citep{patil2023gorilla}.
    The contexts in APIBench consist of many real-world APIs, each of which includes an API name, an API call and an API description.
    The questions in APIBench require to retrieve the API calls based on the given development requirements.
    Due to the ambiguity in the requirements, APIBench serves as the most challenging evaluation task for information seeking.
    We take three sub-tasks from APIBench for evaluations: \textbf{TensorHub (Tens)}, \textbf{TorchHub (Torc)}, and \textbf{Huggingface (Hugg)}.
    In each sub-task, we regard all the candidate APIs as the contexts.
    
\end{itemize}


The \textbf{document-based question-answering (DocQA)} tasks can further reflect how well our DePaC uses the retrieved documents in real-world RAG scenarios.
Specifically, we take three real-world long-document tasks to mimic the process of RAG: given a document-specific question, we provide the model several candidate documents, containing one ground-truth document and other unrelated documents.
The DocQA tasks include:

\begin{itemize}
    \item \textbf{Qasper}~\citep{dasigi2021dataset}. The documents in Qasper are academic research papers and the questions in Qasper are written by NLP practitioners.
    Specifically, after reading only the title and abstract of each paper, the annotators are required to ask an in-depth question which need the information from the full text to get a comprehensive answer. 
    \item \textbf{MultifieldQA}~\citep{bai2023longbench}. The MultifieldQA task aims to test long-document understanding of the model on across diverse fields. The contexts in MultifieldQA are collected from various data sources, including legal documents, government reports, encyclopedias, and academic papers.
    \item  \textbf{NarrativeQA}~\citep{kovcisky2018NarrativeQA}. The NarrativeQA task evaluates how well the model understands the entire long books or movie scripts. 
    Answering the questions in NarrativeQA requires the understanding of the underlying narratives in the given document.
\end{itemize}


For the evaluation metrics, we use exact-match accuracy in the information seeking tasks and F1 score in the DocQA tasks. On information seeking tasks, we set context window number $k$=8 and evenly divide all items into $k$ windows for all PCE approaches. 
On DocQA tasks, we augmented the original QA dataset by expanding the number of documents $k$= 5,10,20 in the context. To avoid exceeding window length when concating documents, we treat each document as a context window for PCE approaches.

\begin{table*}[t]
\renewcommand\arraystretch{1.2}
\caption{Information seeking results.}
\label{tab:ir}
\centering
\resizebox{.75\linewidth}{!}{
\begin{tabular}{lcccccccc}
\toprule
\multirow{2}{*}{Method} & \multirow{2}{*}{FuncNR} & \multirow{2}{*}{EntLR} & \multirow{2}{*}{MVIH} & \multicolumn{3}{c}{APIBench} & \multirow{2}{*}{Avg} \\
& & & & Tens & Torc & Hugg & \\
\midrule
Vanilla~\citep{jiang2023mistral}        & 25.4 & 44.1 & 21.9 & 37.1 & 14.5 & 1.4 & 24.1\\
YaRN~\citep{peng2023YaRN}   & 7.9  & 21.0 & 10.2 & 38.2     & 3.2     & 0.3    & 13.5\\
AVP~\citep{hao2022structured}            & 2.3  & 0.3 & 0.3 & 38.8 & 3.2 & 0.2 & 7.5\\
NBCE~\citep{su2024naive}          & 36.2 & 83.1 & 27.9 & 43.3 & 3.8  & 1.3 & 32.6\\
DePaC (ours)      & \textbf{72.8} & \textbf{87.4} & \textbf{41.6} & \textbf{44.8} & \textbf{16.7} & \textbf{7.5} & \textbf{45.1}\\

\bottomrule
\end{tabular}
}
\end{table*}

\begin{figure*}[t]
    \centering
    \begin{subfigure}[b]{0.49\textwidth}
        \centering
        \includegraphics[width=\textwidth]{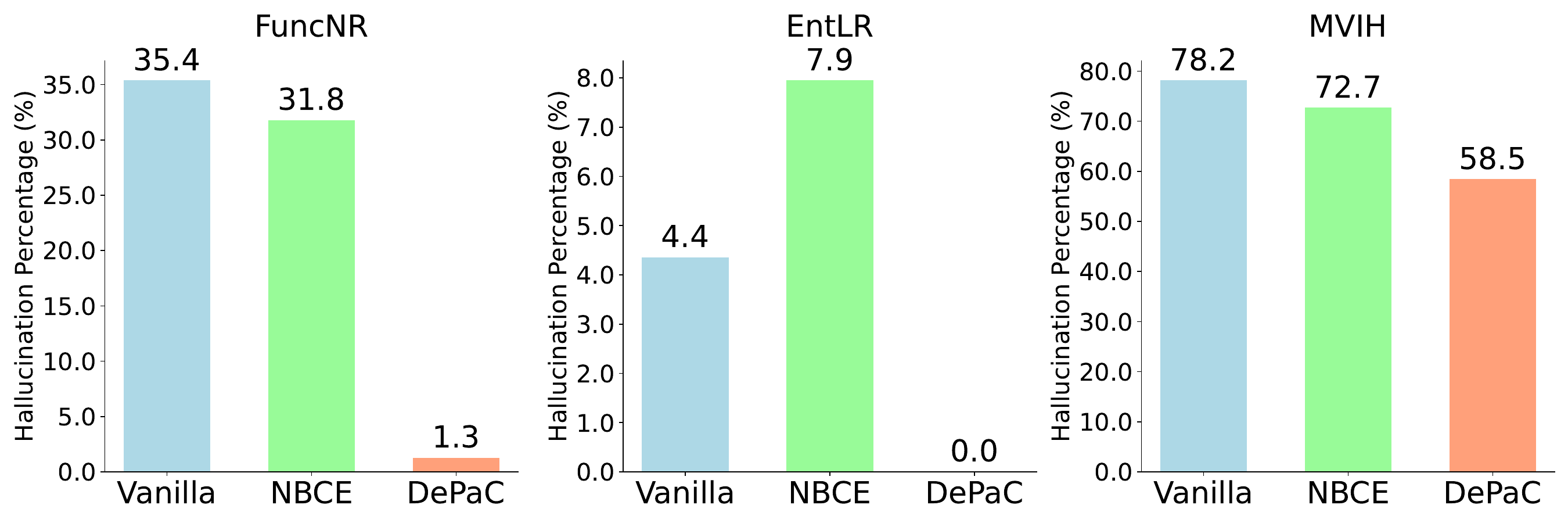}
        \caption{Fact Omission}
        \label{fig:fact_omission}
    \end{subfigure}
    \hfill
    \begin{subfigure}[b]{0.49\textwidth}
        \centering
        \includegraphics[width=\textwidth]{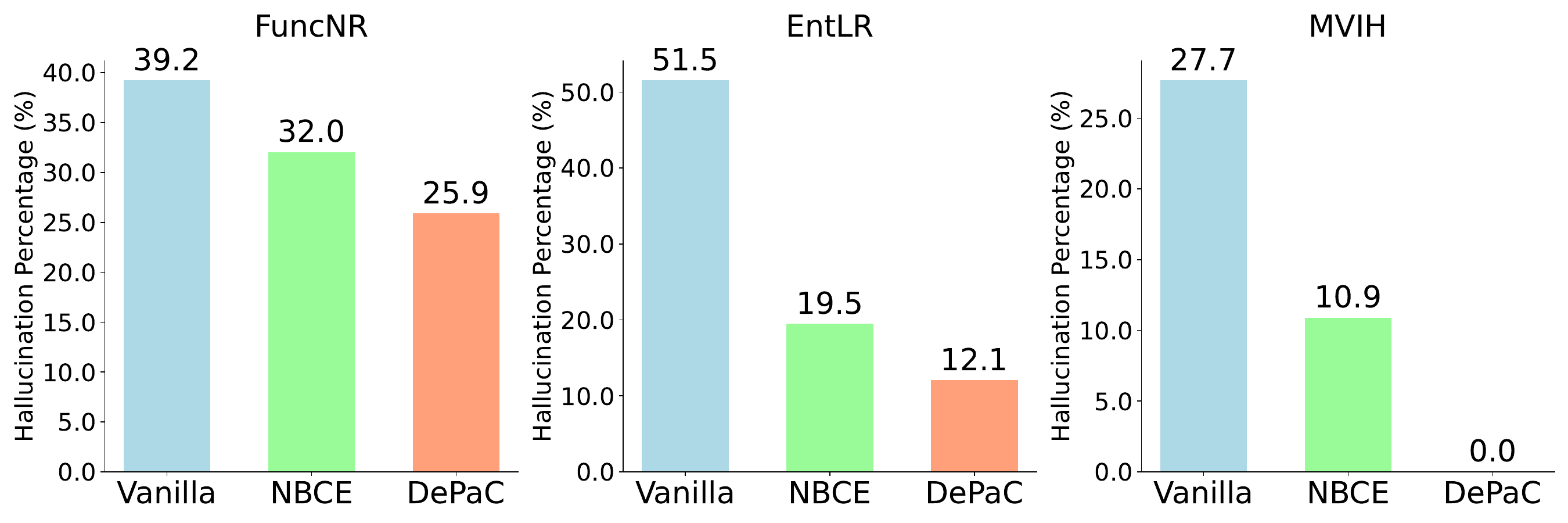}
        \caption{Fact Fabrication}
        \label{fig:fact_fabrication}
    \end{subfigure}
    \caption{Hallucination percentage in responses for the information seeking tasks.}
    \label{fig:fact_hallucination}
\end{figure*}

\subsection{Baselines}
We take Mistral-7B~\citep{jiang2023mistral} as the backbone model for all methods by default.
We mainly compare DePaC with four baseline methods, including the two most effective PCE methods in existing work (i.e., \textbf{AVP}~\citep{hao2022structured, ratner2023parallel} and \textbf{NBCE}~\citep{su2024naive}) and one SOTA-level solution for position extension (i.e., \textbf{YaRN}~\citep{peng2023YaRN}).
\begin{itemize}
    \item \textbf{Vanilla} refers to directly using the vanilla inference approach for a context-limited model~\citep{bai2023longbench}, i.e., concatenating all candidate contexts into input sequence and applying the middle truncation strategy to meet the maximum context length of the model.
    \item \textbf{YaRN}~\citep{peng2023YaRN} is a RoPE-based long context extension approach that expands the context length of Mistral-7B from 32k to 128k.
    With the extended context length, we can feed the full concatenated sequence into the model without truncation.
    \item \textbf{AVP}~\citep{hao2022structured, ratner2023parallel} takes the average aggregation (defined in Equation~\ref{equ:avg}) to aggregate the parallel context windows.
    \item \textbf{NBCE}~\citep{su2024naive} employs the lowest-uncertainty aggregation (defined in Equation~\ref{equ:nbce}) to aggregate the parallel context windows.
\end{itemize}



\subsection{Results and Analysis}


\paragraph{DePaC consistently achieves promising performances across nine tasks.} As shown in Table \ref{tab:ir} and Table \ref{tab:qa}, DePaC achieves better performance than baselines across six information seeking tasks and three DocQA tasks. It is worth noting that although YaRN can process more documents, it performs worse than Vanilla on some tasks. This indicates that long-context extension approach may potentially impair the information seeking capabilities of LLM.

\paragraph{DePaC maintains promising performance with candidate documents number increases.} On DocQA tasks, as the number of documents increases, more redundant information in the context, DePaC still achieves promising performance. DePaC's performance with $k$=20 even surpasses NBCE with $k$=5 (23.9\,vs.\,19.5), further demonstrating DePaC's capability to identify key information from redundant context.

\paragraph{DePaC significantly alleviates fact fabrication and fact omission hallucinations.} We analyze the proportion of hallucinations produced by different approaches on three information seeking tasks (FuncNR, EntLR and MVIH). As shown in Figure~\ref{fig:fact_hallucination}, DePaC significantly reduces the occurrence of both types of hallucinations. DePaC even completely avoids fact omission on EntLR and fact fabrication on MVIH. The detailed hallucination evaluation setup is shown in Appendix \ref{hallucination_setup}.


\begin{figure*}[t]
    \centering
    \includegraphics[width=.85\textwidth]{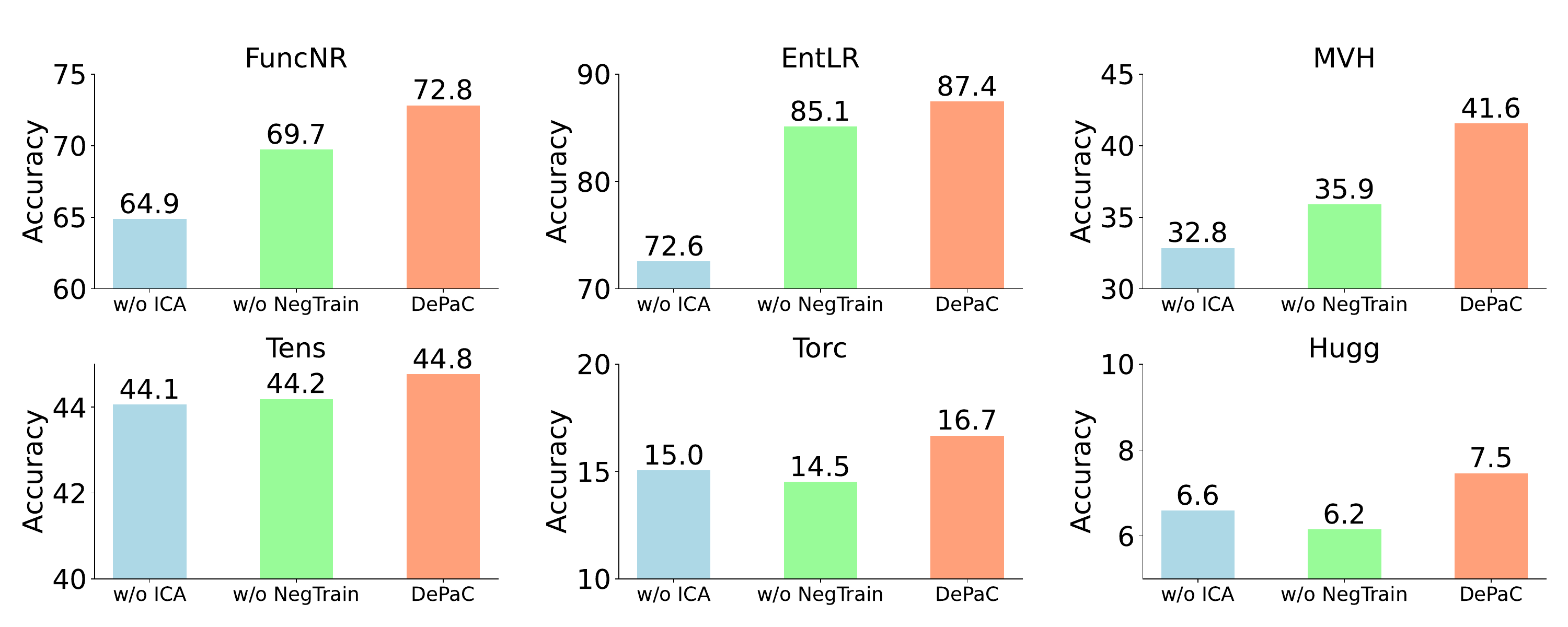}
    \caption{
    Performance of DePaC without NegTrain or ICA. w/o NegTrain refers to DePaC with positive training, while w/o ICA refers to replace ICA with lowest-uncertainty aggregation of NBCE.
    }
    \label{fig:ablation}
\end{figure*}

\paragraph{Both information-calibrated aggregation and context-aware negative training are essential for DePaC performance.} 
We compare DePaC with two ablation setting: 
(1) \textbf{DePaC w/o NegTrain.} We reconstruct a \textbf{Pos}itive \textbf{Train}ing \textbf{(PosTrain)} dataset composed solely of positive samples, with the sample size as NegTrain dataset, and finetune Mistral-7B with PosTrain dataset.
(2) \textbf{DePaC w/o ICA.} We only replace the information-calibrated aggregation function of DePaC with lowest-uncertainty aggregation.
We conducte ablation study on the six information seeking datasets. As shown in Figure \ref{fig:ablation}, the ablation results indicate that both parts of DePaC are essential for its performance.




\paragraph{DePaC also achieves promising performance on different LLMs.} We also evaluate DePaC on another advanced LLM, Llama3-8B\footnote{https://huggingface.co/meta-llama/Meta-Llama-3-8B-Instruct}. Table \ref{tab:qa_llama} shows the different PCE performance with Llama3-8B on the three QA datasets, DePaC also achieves promising performance with Llama3-8B model. It is worth mentioning that, while the performance of vanilla inference with Llama3-8B is significantly lower than with Mistral-7B (7.7\,vs.\,21.6), DePaC with Llama3-8B achieves comparable performance to Mistral-7B (24.2\,vs.\,24.8). This further confirms our finding that DePaC can effectively leverage the short-context processing capabilities of LLMs to extract useful information from long-context.


\section{Related Work} \label{sec:related}

\paragraph{Retrieval-Augmented Generation (RAG) for LLM.} To address hallucination issue of LLM, Retrieval-augmented generation~\citep{lewis2020retrieval, gao2023retrieval, cheng2024lift, asai2023self} has been applied in many fields, including question answering~\citep{zhang2024raft}, code generation~\citep{zhou2022docprompting,ma2024compositional} and recommendation~\citep{zeng2024federated}. The performance of RAG is limited by the effectiveness of retriever and the information utilization capability of LLM. Some work focus on enhancing the retriever's capabilities~\citep{wang2023improving, lewis2020retrieval}. ~\citet{shi2024compressing} compresses the retrieved information for LLM. Some work proposes iterative RAG~\citep{jiang2023active,shao2023enhancing, cheng2024lift} to help the model progressively utilize document information.
Some work~\citep{asai2023self, dhuliawala2023chain, feng2024don} utilizes prompt engineer to aggregate information from multiple documents to generate a final answer. These methods often lead to information omission during the aggregation process. In this work, we utilize PCE to directly aggregate information from multiple documents when predicting the next token, enhance the accuracy and efficiency of information utilization.

\paragraph{LLM with Parallel Context Extension (PCE).} Recent research has proposed some PCE approaches to aggregate multiple context windows into a unified representation space, extending context length of LLM. Some research ~\citep{hao2022structured,ratner2023parallel, li2024paraicl} aggregates by average aggregation mechanisms. ~\citet{su2024naive} proposes NBCE to aggregates by lowest-uncertainty aggregation mechanisms. Previous PCE work primarily focuses on increasing in-context learning examples, and faces hallucination issues when applied for RAG~\citep{yang2023revisiting}. Beyond parallel context extension for existing LLM, \citet{yen2024long} also proposes encoder-decoder architecture to implement parallel context. In this work, we propose DePaC to alleviate the hallucination issues of PCE for RAG scenarios. To the best of our knowledge, we are the first work to apply PCE to RAG scenarios.


\section{Conclusion} \label{sec:conclusion}
In this paper, we propose DePaC to address two types of in-context hallucination issues of parallel context extension on RAG. DePaC consists of two key components: (1) a context-aware negative training technique to mitigate fact fabrication, and (2) an information-calibrated aggregation method to address fact omission issue. Both experiments on information seeking and DocQA tasks show the effectiveness of DePaC.

\section{Limitations} \label{sec:limitation}

\textbf{Data generation cost.} We rely on GPT-4-Turbo to generate our training data, which cost around 90\$ for API calling. Future work should attempt to generate data using cheaper models without compromising data quality.

\textbf{Training cost.} Our training process consumes some computational resources, but it's a one-time effort. Given the advantages of our method in terms of inference efficiency and accuracy, we believe these offline costs are justified.

\section*{Acknowledgments}
This work is supported by National Key Research and Development Program of China (Grant No. 2023YFB4503803).






\bibliography{reference}

\begin{thebibliography}{41}
\providecommand{\natexlab}[1]{#1}

\bibitem[{An et~al.(2024)An, Ma, Lin, Zheng, and Lou}]{an2024make}
Shengnan An, Zexiong Ma, Zeqi Lin, Nanning Zheng, and Jian-Guang Lou. 2024.
\newblock \href {https://arxiv.org/abs/2404.16811} {Make your llm fully utilize the context}.
\newblock \emph{Preprint}, arXiv:2404.16811.

\bibitem[{Asai et~al.(2023)Asai, Wu, Wang, Sil, and Hajishirzi}]{asai2023self}
Akari Asai, Zeqiu Wu, Yizhong Wang, Avirup Sil, and Hannaneh Hajishirzi. 2023.
\newblock Self-rag: Learning to retrieve, generate, and critique through self-reflection.
\newblock \emph{arXiv preprint arXiv:2310.11511}.

\bibitem[{Bai et~al.(2023)Bai, Lv, Zhang, Lyu, Tang, Huang, Du, Liu, Zeng, Hou et~al.}]{bai2023longbench}
Yushi Bai, Xin Lv, Jiajie Zhang, Hongchang Lyu, Jiankai Tang, Zhidian Huang, Zhengxiao Du, Xiao Liu, Aohan Zeng, Lei Hou, et~al. 2023.
\newblock Longbench: A bilingual, multitask benchmark for long context understanding.
\newblock \emph{arXiv preprint arXiv:2308.14508}.

\bibitem[{Chen et~al.(2024)Chen, Lin, Han, and Sun}]{chen2024benchmarking}
Jiawei Chen, Hongyu Lin, Xianpei Han, and Le~Sun. 2024.
\newblock Benchmarking large language models in retrieval-augmented generation.
\newblock In \emph{Proceedings of the AAAI Conference on Artificial Intelligence}, volume~38, pages 17754--17762.

\bibitem[{Cheng et~al.(2024)Cheng, Luo, Chen, Liu, Zhao, and Yan}]{cheng2024lift}
Xin Cheng, Di~Luo, Xiuying Chen, Lemao Liu, Dongyan Zhao, and Rui Yan. 2024.
\newblock Lift yourself up: Retrieval-augmented text generation with self-memory.
\newblock \emph{Advances in Neural Information Processing Systems}, 36.

\bibitem[{Dao(2023)}]{dao2023flashattention2}
Tri Dao. 2023.
\newblock Flash{A}ttention-2: Faster attention with better parallelism and work partitioning.

\bibitem[{Dasigi et~al.(2021)Dasigi, Lo, Beltagy, Cohan, Smith, and Gardner}]{dasigi2021dataset}
Pradeep Dasigi, Kyle Lo, Iz~Beltagy, Arman Cohan, Noah~A Smith, and Matt Gardner. 2021.
\newblock A dataset of information-seeking questions and answers anchored in research papers.
\newblock In \emph{Proceedings of the 2021 Conference of the North American Chapter of the Association for Computational Linguistics: Human Language Technologies}, pages 4599--4610.

\bibitem[{Dhuliawala et~al.(2023)Dhuliawala, Komeili, Xu, Raileanu, Li, Celikyilmaz, and Weston}]{dhuliawala2023chain}
Shehzaad Dhuliawala, Mojtaba Komeili, Jing Xu, Roberta Raileanu, Xian Li, Asli Celikyilmaz, and Jason Weston. 2023.
\newblock Chain-of-verification reduces hallucination in large language models.
\newblock \emph{arXiv preprint arXiv:2309.11495}.

\bibitem[{Feng et~al.(2024)Feng, Shi, Wang, Ding, Balachandran, and Tsvetkov}]{feng2024don}
Shangbin Feng, Weijia Shi, Yike Wang, Wenxuan Ding, Vidhisha Balachandran, and Yulia Tsvetkov. 2024.
\newblock Don't hallucinate, abstain: Identifying llm knowledge gaps via multi-llm collaboration.
\newblock \emph{arXiv preprint arXiv:2402.00367}.

\bibitem[{Gao et~al.(2023)Gao, Xiong, Gao, Jia, Pan, Bi, Dai, Sun, and Wang}]{gao2023retrieval}
Yunfan Gao, Yun Xiong, Xinyu Gao, Kangxiang Jia, Jinliu Pan, Yuxi Bi, Yi~Dai, Jiawei Sun, and Haofen Wang. 2023.
\newblock Retrieval-augmented generation for large language models: A survey.
\newblock \emph{arXiv preprint arXiv:2312.10997}.

\bibitem[{Ghoshal and Tucker(2022)}]{ghoshal2022calibrated}
Biraja Ghoshal and Allan Tucker. 2022.
\newblock On calibrated model uncertainty in deep learning.
\newblock \emph{arXiv preprint arXiv:2206.07795}.

\bibitem[{Hao et~al.(2022)Hao, Sun, Dong, Han, Gu, and Wei}]{hao2022structured}
Yaru Hao, Yutao Sun, Li~Dong, Zhixiong Han, Yuxian Gu, and Furu Wei. 2022.
\newblock Structured prompting: Scaling in-context learning to 1,000 examples.
\newblock \emph{arXiv preprint arXiv:2212.06713}.

\bibitem[{Hsieh et~al.(2024)Hsieh, Sun, Kriman, Acharya, Rekesh, Jia, and Ginsburg}]{hsieh2024ruler}
Cheng-Ping Hsieh, Simeng Sun, Samuel Kriman, Shantanu Acharya, Dima Rekesh, Fei Jia, and Boris Ginsburg. 2024.
\newblock \href {https://arxiv.org/abs/2404.06654} {Ruler: What's the real context size of your long-context language models?}
\newblock \emph{Preprint}, arXiv:2404.06654.

\bibitem[{Ji et~al.(2023)Ji, Lee, Frieske, Yu, Su, Xu, Ishii, Bang, Madotto, and Fung}]{ji2023survey}
Ziwei Ji, Nayeon Lee, Rita Frieske, Tiezheng Yu, Dan Su, Yan Xu, Etsuko Ishii, Ye~Jin Bang, Andrea Madotto, and Pascale Fung. 2023.
\newblock Survey of hallucination in natural language generation.
\newblock \emph{ACM Computing Surveys}, 55(12):1--38.

\bibitem[{Jiang et~al.(2023{\natexlab{a}})Jiang, Sablayrolles, Mensch, Bamford, Chaplot, Casas, Bressand, Lengyel, Lample, Saulnier et~al.}]{jiang2023mistral}
Albert~Q Jiang, Alexandre Sablayrolles, Arthur Mensch, Chris Bamford, Devendra~Singh Chaplot, Diego de~las Casas, Florian Bressand, Gianna Lengyel, Guillaume Lample, Lucile Saulnier, et~al. 2023{\natexlab{a}}.
\newblock Mistral 7b.
\newblock \emph{arXiv preprint arXiv:2310.06825}.

\bibitem[{Jiang et~al.(2023{\natexlab{b}})Jiang, Xu, Gao, Sun, Liu, Dwivedi-Yu, Yang, Callan, and Neubig}]{jiang2023active}
Zhengbao Jiang, Frank~F Xu, Luyu Gao, Zhiqing Sun, Qian Liu, Jane Dwivedi-Yu, Yiming Yang, Jamie Callan, and Graham Neubig. 2023{\natexlab{b}}.
\newblock Active retrieval augmented generation.
\newblock \emph{arXiv preprint arXiv:2305.06983}.

\bibitem[{Joshi et~al.(2017)Joshi, Choi, Weld, and Zettlemoyer}]{joshi2017triviaqa}
Mandar Joshi, Eunsol Choi, Daniel~S Weld, and Luke Zettlemoyer. 2017.
\newblock Triviaqa: A large scale distantly supervised challenge dataset for reading comprehension.
\newblock \emph{arXiv preprint arXiv:1705.03551}.

\bibitem[{Ko{\v{c}}isk{\`y} et~al.(2018)Ko{\v{c}}isk{\`y}, Schwarz, Blunsom, Dyer, Hermann, Melis, and Grefenstette}]{kovcisky2018NarrativeQA}
Tom{\'a}{\v{s}} Ko{\v{c}}isk{\`y}, Jonathan Schwarz, Phil Blunsom, Chris Dyer, Karl~Moritz Hermann, G{\'a}bor Melis, and Edward Grefenstette. 2018.
\newblock The narrativeqa reading comprehension challenge.
\newblock \emph{Transactions of the Association for Computational Linguistics}, 6:317--328.

\bibitem[{Kullback and Leibler(1951)}]{10.1214/aoms/1177729694}
S.~Kullback and R.~A. Leibler. 1951.
\newblock \href {https://doi.org/10.1214/aoms/1177729694} {{On Information and Sufficiency}}.
\newblock \emph{The Annals of Mathematical Statistics}, 22(1):79 -- 86.

\bibitem[{Kwiatkowski et~al.(2019)Kwiatkowski, Palomaki, Redfield, Collins, Parikh, Alberti, Epstein, Polosukhin, Devlin, Lee et~al.}]{kwiatkowski2019natural}
Tom Kwiatkowski, Jennimaria Palomaki, Olivia Redfield, Michael Collins, Ankur Parikh, Chris Alberti, Danielle Epstein, Illia Polosukhin, Jacob Devlin, Kenton Lee, et~al. 2019.
\newblock Natural questions: a benchmark for question answering research.
\newblock \emph{Transactions of the Association for Computational Linguistics}, 7:453--466.

\bibitem[{Lewis et~al.(2020)Lewis, Perez, Piktus, Petroni, Karpukhin, Goyal, K{\"u}ttler, Lewis, Yih, Rockt{\"a}schel et~al.}]{lewis2020retrieval}
Patrick Lewis, Ethan Perez, Aleksandra Piktus, Fabio Petroni, Vladimir Karpukhin, Naman Goyal, Heinrich K{\"u}ttler, Mike Lewis, Wen-tau Yih, Tim Rockt{\"a}schel, et~al. 2020.
\newblock Retrieval-augmented generation for knowledge-intensive nlp tasks.
\newblock \emph{Advances in Neural Information Processing Systems}, 33:9459--9474.

\bibitem[{Li et~al.(2023)Li, Zi, Muennighoff, Kocetkov, Mou, Marone, Akiki, Jia, Chim, Liu et~al.}]{li2023starcoder}
Raymond Li, Yangtian Zi, Niklas Muennighoff, Denis Kocetkov, Chenghao Mou, Marc Marone, Christopher Akiki, LI~Jia, Jenny Chim, Qian Liu, et~al. 2023.
\newblock Starcoder: may the source be with you!
\newblock \emph{Transactions on Machine Learning Research}.

\bibitem[{Li et~al.(2024)Li, Nguyen, Joty, and Bing}]{li2024paraicl}
Xingxuan Li, Xuan-Phi Nguyen, Shafiq Joty, and Lidong Bing. 2024.
\newblock Paraicl: Towards robust parallel in-context learning.
\newblock \emph{arXiv preprint arXiv:2404.00570}.

\bibitem[{Ma et~al.(2024)Ma, An, Xie, and Lin}]{ma2024compositional}
Zexiong Ma, Shengnan An, Bing Xie, and Zeqi Lin. 2024.
\newblock Compositional api recommendation for library-oriented code generation.
\newblock \emph{arXiv preprint arXiv:2402.19431}.

\bibitem[{OpenAI(2023)}]{openai2023gpt4}
OpenAI. 2023.
\newblock \href {https://arxiv.org/abs/2303.08774} {Gpt-4 technical report}.
\newblock \emph{Preprint}, arXiv:2303.08774.

\bibitem[{Patil et~al.(2023)Patil, Zhang, Wang, and Gonzalez}]{patil2023gorilla}
Shishir~G Patil, Tianjun Zhang, Xin Wang, and Joseph~E Gonzalez. 2023.
\newblock Gorilla: Large language model connected with massive apis.
\newblock \emph{arXiv preprint arXiv:2305.15334}.

\bibitem[{Peng et~al.(2023)Peng, Quesnelle, Fan, and Shippole}]{peng2023YaRN}
Bowen Peng, Jeffrey Quesnelle, Honglu Fan, and Enrico Shippole. 2023.
\newblock Yarn: Efficient context window extension of large language models.
\newblock In \emph{The Twelfth International Conference on Learning Representations}.

\bibitem[{Raffel et~al.(2020)Raffel, Shazeer, Roberts, Lee, Narang, Matena, Zhou, Li, and Liu}]{raffel2020exploring}
Colin Raffel, Noam Shazeer, Adam Roberts, Katherine Lee, Sharan Narang, Michael Matena, Yanqi Zhou, Wei Li, and Peter~J Liu. 2020.
\newblock Exploring the limits of transfer learning with a unified text-to-text transformer.
\newblock \emph{Journal of machine learning research}, 21(140):1--67.

\bibitem[{Ratner et~al.(2023)Ratner, Levine, Belinkov, Ram, Magar, Abend, Karpas, Shashua, Leyton-Brown, and Shoham}]{ratner2023parallel}
Nir Ratner, Yoav Levine, Yonatan Belinkov, Ori Ram, Inbal Magar, Omri Abend, Ehud Karpas, Amnon Shashua, Kevin Leyton-Brown, and Yoav Shoham. 2023.
\newblock Parallel context windows for large language models.
\newblock In \emph{Proceedings of the 61st Annual Meeting of the Association for Computational Linguistics (Volume 1: Long Papers)}, pages 6383--6402.

\bibitem[{Rawte et~al.(2023)Rawte, Sheth, and Das}]{rawte2023survey}
Vipula Rawte, Amit Sheth, and Amitava Das. 2023.
\newblock A survey of hallucination in large foundation models.
\newblock \emph{arXiv preprint arXiv:2309.05922}.

\bibitem[{Shao et~al.(2023)Shao, Gong, Shen, Huang, Duan, and Chen}]{shao2023enhancing}
Zhihong Shao, Yeyun Gong, Yelong Shen, Minlie Huang, Nan Duan, and Weizhu Chen. 2023.
\newblock Enhancing retrieval-augmented large language models with iterative retrieval-generation synergy.
\newblock \emph{arXiv preprint arXiv:2305.15294}.

\bibitem[{Shi et~al.(2024)Shi, Sun, Li, and Xu}]{shi2024compressing}
Kaize Shi, Xueyao Sun, Qing Li, and Guandong Xu. 2024.
\newblock Compressing long context for enhancing rag with amr-based concept distillation.
\newblock \emph{arXiv preprint arXiv:2405.03085}.

\bibitem[{Su et~al.(2024)Su, Ahmed, Ao, Zhu, Liu et~al.}]{su2024naive}
Jianlin Su, Murtadha Ahmed, Luo Ao, Mingren Zhu, Yunfeng Liu, et~al. 2024.
\newblock Naive bayes-based context extension for large language models.
\newblock \emph{arXiv preprint arXiv:2403.17552}.

\bibitem[{Touvron et~al.(2023)Touvron, Martin, Stone, Albert, Almahairi, Babaei, Bashlykov, Batra, Bhargava, Bhosale et~al.}]{touvron2023llama}
Hugo Touvron, Louis Martin, Kevin Stone, Peter Albert, Amjad Almahairi, Yasmine Babaei, Nikolay Bashlykov, Soumya Batra, Prajjwal Bhargava, Shruti Bhosale, et~al. 2023.
\newblock Llama 2: Open foundation and fine-tuned chat models.
\newblock \emph{arXiv preprint arXiv:2307.09288}.

\bibitem[{Wang et~al.(2023)Wang, Yang, Huang, Yang, Majumder, and Wei}]{wang2023improving}
Liang Wang, Nan Yang, Xiaolong Huang, Linjun Yang, Rangan Majumder, and Furu Wei. 2023.
\newblock Improving text embeddings with large language models.
\newblock \emph{arXiv preprint arXiv:2401.00368}.

\bibitem[{Weng(2024)}]{extrinsic_hallucination}
Lilian Weng. 2024.
\newblock \href {https://lilianweng.github.io/posts/2024-07-07-hallucination/} {Extrinsic hallucinations in llms}.

\bibitem[{Yang et~al.(2023)Yang, Liu, Men, Zeng, Dong, and Tang}]{yang2023revisiting}
Kejuan Yang, Xiao Liu, Kaiwen Men, Aohan Zeng, Yuxiao Dong, and Jie Tang. 2023.
\newblock Revisiting parallel context windows: A frustratingly simple alternative and chain-of-thought deterioration.
\newblock \emph{arXiv preprint arXiv:2305.15262}.

\bibitem[{Yen et~al.(2024)Yen, Gao, and Chen}]{yen2024long}
Howard Yen, Tianyu Gao, and Danqi Chen. 2024.
\newblock Long-context language modeling with parallel context encoding.
\newblock \emph{arXiv preprint arXiv:2402.16617}.

\bibitem[{Zeng et~al.(2024)Zeng, Yue, Jiang, and Wang}]{zeng2024federated}
Huimin Zeng, Zhenrui Yue, Qian Jiang, and Dong Wang. 2024.
\newblock Federated recommendation via hybrid retrieval augmented generation.
\newblock \emph{arXiv preprint arXiv:2403.04256}.

\bibitem[{Zhang et~al.(2024)Zhang, Patil, Jain, Shen, Zaharia, Stoica, and Gonzalez}]{zhang2024raft}
Tianjun Zhang, Shishir~G Patil, Naman Jain, Sheng Shen, Matei Zaharia, Ion Stoica, and Joseph~E Gonzalez. 2024.
\newblock Raft: Adapting language model to domain specific rag.
\newblock \emph{arXiv preprint arXiv:2403.10131}.

\bibitem[{Zhou et~al.(2022)Zhou, Alon, Xu, Wang, Jiang, and Neubig}]{zhou2022docprompting}
Shuyan Zhou, Uri Alon, Frank~F Xu, Zhiruo Wang, Zhengbao Jiang, and Graham Neubig. 2022.
\newblock Docprompting: Generating code by retrieving the docs.
\newblock \emph{arXiv preprint arXiv:2207.05987}.

\end{thebibliography}

\appendix
\clearpage
This is the Appendix of the paper: \textit{Dehallucinating Parallel Context Extension for Retrieval-Augmented Generation}.

\section{More Formula Details}
The Kullback-Leibler (KL) divergence for discrete probability distributions \(\mathbf{P_1}\) and \(\mathbf{P_2}\) is defined as:

\begin{equation}
\mathrm{D_{KL}}(\mathbf{P_1}\ ||\ \mathbf{P_2}) = \sum_{i} \mathbf{P_1}(i) \log \frac{\mathbf{P_1}(i)}{\mathbf{P_2}(i)}
\end{equation}


The cross-entropy loss function is defined as:


\begin{flalign}
   &\mathrm{CE}[p_{\theta}(\ \cdot\ |\ d_{j} \oplus \mathcal{Q}),\ \mathcal{A}] =  &&\\\nonumber
   & -\sum_{i=1}^{n} \log p_{\theta}(\mathcal{A}_i\ |\ d_{j} \oplus \mathcal{Q} \oplus \mathcal{A}_{1:i-1}) &&
\end{flalign}

where $\mathcal{A}_i$ is the $i$-th token in g round-truth answers, $n$ is the sequence length of ground-truth.
$p_{\theta}(\mathcal{A}_i | d_{j} \oplus \mathcal{Q} \oplus \mathcal{A}_{1:i-1})$ is the probability of generating $\mathcal{A}_i$ given the input $d_{j} \oplus \mathcal{Q} \oplus \mathcal{A}_{1:i-1}$.

\section{DePaC Simplified Form}\label{sec:simplified_form}

Notice that one implicate constraint in Equation~\ref{equ:ours_ori_p} is $\gamma \gg C(\mathbf{p_{i, j}}, \mathbf{p_{i, c}})$ as we hope to directly filter out irrelevant context windows.
To simplify this constraint for implementation, we rewrite Equation~\ref{equ:ours_ori_p} as the product of two terms and modify Equation~\ref{equ:ours_ori_c} to make sure $\hat{C}(\mathbf{p_{i, j}}, \mathbf{p_{i, c}})\geq0$,
\begin{flalign}\label{equ:ours_new_p}
    &\mathbf{p_{i}} = &&\\\nonumber
    &\argmax_{\mathbf{p_{i, j}}} \hat{C}(\mathbf{p_{i, j}}, \mathbf{p_{i, c}}) \cdot \mathbb{I}(\argmax_{k}\mathbf{p_{i, j}}^{k}=t_{d}),&&
\end{flalign}
\begin{equation}\label{equ:ours_new_c}
    \hat{C}(\mathbf{p_{i, j}}, \mathbf{p_{i, c}}) = \max_{k}\mathbf{p_{i, j}}^{k} + \beta\cdot\Delta(\mathbf{p_{i, j}}, \mathbf{p_{i, c}}),\quad
\end{equation}
where we use $\max_{k}\mathbf{p_{i, j}}^{k}$ to estimate the output certainty, and $\beta > 0$ is hyper-parameter.
For the output of deep learning models, a higher $\max_{k}\mathbf{p_{i, j}}^{k}$ always indicates a higher certainty in practice~\citep{ghoshal2022calibrated}.
We set $\beta=0.2$ by default and analyze the choice of $\beta$ in Appendix~\ref{sec:beta_ablation}.

\section{Hyperparameter Settings}\label{sec:beta_ablation}
\begin{figure}[h]
\centering
\includegraphics[width=0.7\linewidth]{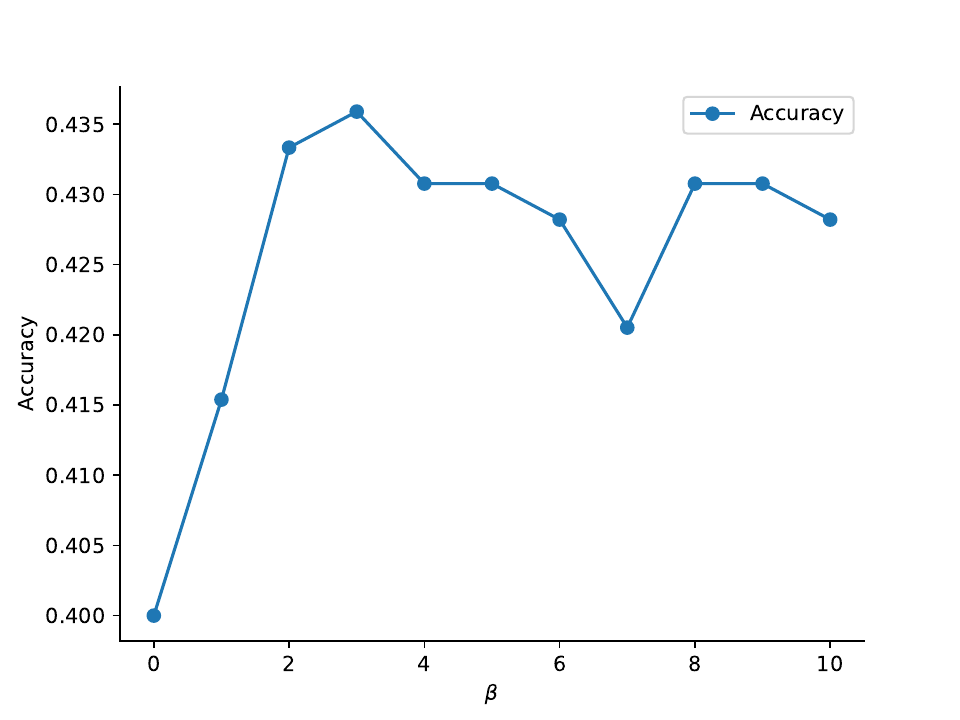}
\vspace{-3mm}
\caption{DePaC performance with different $beta$}
\label{fig:beta_ablation}
\end{figure}

We conducted $\beta$ ablation study on the EntLR dataset. The result in Figure \ref{fig:beta_ablation} indicates that $\beta \in [0.2, 0.3]$ achieves better trade-off between information entropy and KL divergence. We set $\beta=0.2$ in our experiments.

\begin{figure}[t]
    \centering
    \includegraphics[width=.8\linewidth]{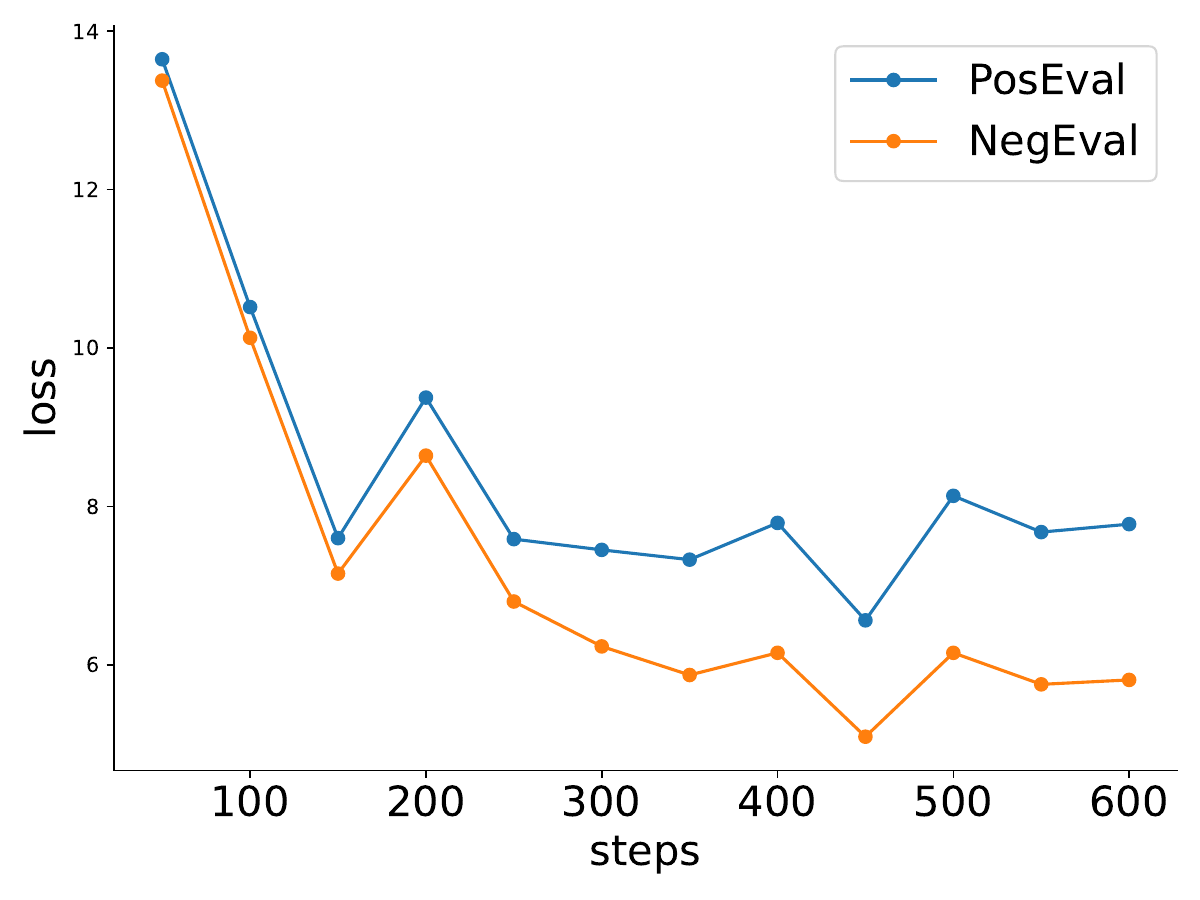}
    \caption{Rejection token prediction loss on PosEval and NegEval over context-aware negative training steps.}
    \label{fig:ppl_over_steps}
\end{figure}
\section{Analysis on NegTrain}
\paragraph{Context-aware Negative training can improve the ability of refusing to answer questions with unrelated documents.} 
We constructed an additional 4.4K positive samples (PosEval) and negative samples (NegEval), using the same data construction method as NegTrain, but with different seed documents. PosEval represents the situation that documents are related to the question, while NegEval represents the opposite.
We compare the rejection token $t_d$ prediction loss on PosEval and NegEval datasets with different NegTrain steps. Figure \ref{fig:ppl_over_steps} shows that NegTrain can increase the probability difference between refusing to answer questions with unrelated document and related document. 

\begin{figure*}[h]
    \centering
    \includegraphics[width=.99\textwidth]{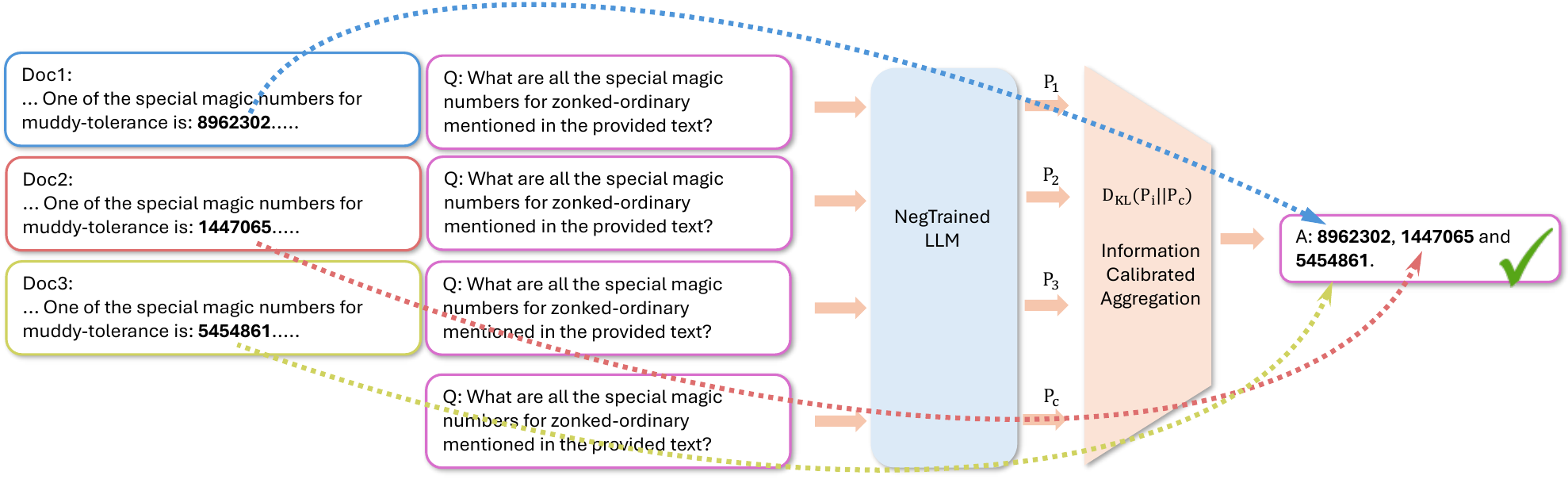}
    \caption{
    DePaC can switch context window for multi-hop questions.
    }
    \label{fig:switch}
\end{figure*}

\begin{table}[h]
\renewcommand\arraystretch{1.2}
\caption{Comparison results between DePaC and aggregation approaches for RAG.}
\label{tab:comparison_with_aggregation}
\centering
\resizebox{\linewidth}{!}{
\begin{tabular}{lccc}
\toprule
Method & NaturalQuestions & TriviaQA & RGB\\
\midrule
SelfRAG~\citep{asai2023self} & 28.67 & 74.33 & 75.33 \\
CoVe~\citep{dhuliawala2023chain} & 26.67 & 68.67 & 76.33 \\
COMPETE~\citep{feng2024don} & 22.67 & 69.00 & 74.00 \\
DePaC (ours) & \textbf{33.67} & \textbf{88.33} & \textbf{94.33} \\
\bottomrule
\end{tabular}
}
\end{table}

\begin{table}[h]
\renewcommand\arraystretch{1.2}
\caption{Multi-hop DocQA results.}
\label{tab:multi_hop_docqa}
\centering
\resizebox{\linewidth}{!}{
\begin{tabular}{lccc}
\toprule
Model & Method & 2WikimQA & HotPotQA\\
\midrule
Llama2-13B & Vanilla~\citep{jiang2023mistral} & 12.44 & 11.25 \\
Llama2-13B & NBCE~\citep{su2024naive} & 22.28 & 19.26 \\
Llama2-13B & DePaC (ours) & \textbf{29.09} & \textbf{23.38} \\
\midrule
Mistral-7B & Vanilla~\citep{jiang2023mistral} & 19.04 & 12.01 \\
Mistral-7B & NBCE~\citep{su2024naive} & 17.45 & 10.52 \\
Mistral-7B & DePaC (ours) & \textbf{29.72} & \textbf{30.95} \\
\bottomrule
\end{tabular}
}
\end{table}

\section{More Evaluation Results}

\paragraph{DePaC performs better than aggregation approaches for RAG.} We also compare DePaC with previous aggregation approaches specific to RAG ~\citep{asai2023self} or can be applied to RAG ~\citep{dhuliawala2023chain,feng2024don}, the results in Table \ref{tab:comparison_with_aggregation} show that DePaC outperforms other aggregation approaches on different datasets ~\citep{kwiatkowski2019natural, joshi2017triviaqa, chen2024benchmarking}.

\paragraph{DePaC with COT maintains performance advantage on multi-hop DocQA across different LLMs.} We evaluate on 2WikimQA and HotPotQA datasets using Mistral-7B and Llama2-13B as base-models. The results in Table \ref{tab:multi_hop_docqa} show that DePaC with different models still maintains its performance advantage on multi-hop QA datasets. We make the prompt for multi-hop QA datasets end with \textit{"Let's think step by step, "}, this helps DePaC first seeks useful information across different contexts before generate the final answer.

    
    


    
    


\section{Case Study}
\paragraph{DePaC can switch context window for multi-hop questions.} Some questions require synthesizing information from multiple documents, which demands DePaC to switch between different context windows during the generation process. As shown in Table \ref{tab:ir}, results on the MVIH dataset indicate that DePaC can achieve better performance on such tasks. As shown in Figure \ref{fig:switch}, DePaC can find all magic numbers located in different documents.

\begin{figure}[h]
\centering
\includegraphics[width=0.7\linewidth]{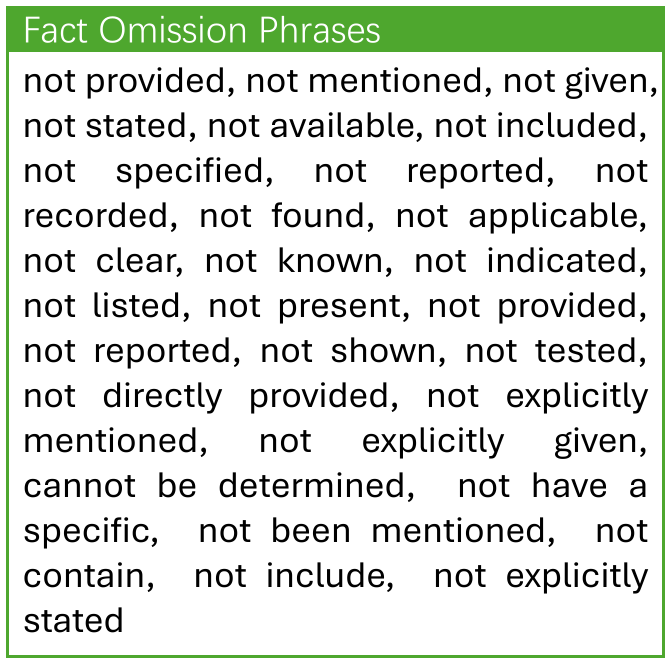}
\vspace{-3mm}
\caption{Fact omission phrases.}
\label{fig:fact_omission_phrases}
\end{figure}

\section{Hallucination Definition and Evaluation Setup}\label{hallucination_setup}
Previous work~\citep{extrinsic_hallucination} categorizes hallucination into two types:
(1) \textbf{extrinsic hallucination}, where the output of LLM is not grounded by the pre-training dataset or external world knowledge.
(2) \textbf{in-context hallucination}, where the output of the model is inconsistent with the source content in context.
In this work we focus on two types of in-context hallucination:
(1) \textbf{fact fabrication}, where LLMs present claims that are not supported by the contexts.
(2) \textbf{fact omission}, where LLMs fail to present claims that are supported by the contexts.

We done in-context hallucination evaluation on three information seeking tasks (FuncNR, EntLR and MVIH), as they are evaluated by exact-match score, makes them easier to analyze than QA tasks. Since these tasks have clear answers in the document and all incorrect outputs are hallucinations, we manually analyzed the data to define 27 fact omission phrases (shown in Figure \ref{fig:fact_omission_phrases}), counted the incorrect outputs that appeared with these phrases as fact omission, and classified other errors as fact fabrication.

\begin{figure}[t]
\centering
\includegraphics[width=.7\linewidth]{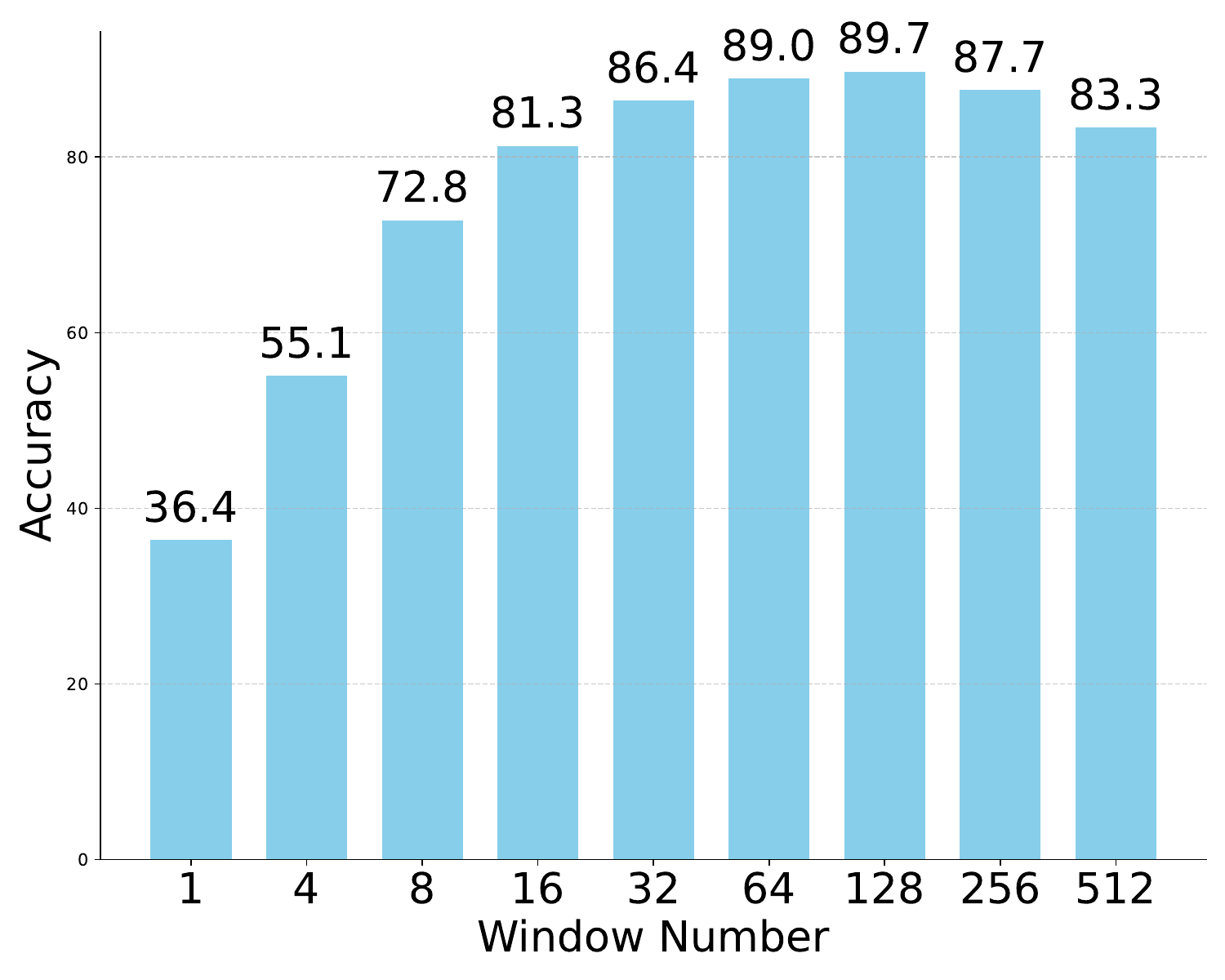}
\caption{DePaC performance at different degrees of context window parallelism.}
\label{fig:shorter_window}
\end{figure}

\section{Window Number Analysis} 
To analyze DePaC's performance with different numbers of windows, we conduct experiments on the FuncNR dataset, keeping the total number of candidate functions constant while varying the number of windows into which the context is divided. The results in Figure \ref{fig:shorter_window} show that as the number of windows increases (form 4 to 128), DePaC's information-seeking ability improves; however, when the number of windows becomes too large (larger than 256), there may be a slight performance decline. All DePaC with split-window outperforms the single-window, further validating the effectiveness of DePaC with parallel context windows.


\section{Broader Impacts}\label{sec:impact}
This work used GPT-4-Turbo to generate training data. Therefore, our fine-tuned model may inherit the potential risks of GPT-4-Turbo in terms of ethical and safety issues.

\end{document}